\documentclass[conference]{IEEEtran}

\usepackage{graphicx}
\usepackage{color}

\usepackage{url}

\begin{document}
\title{Memory-Centred Cognitive Architectures for \\Robots Interacting Socially with Humans}

\author{\IEEEauthorblockN{Paul Baxter}
\IEEEauthorblockA{Centre for Robotics and Neural Systems\\
The Cognition Institute\\
Plymouth University, U.K.\\
\url{paul.baxter@plymouth.ac.uk}}
}

\maketitle

\begin{abstract}

%Memory-centred cognition is good, and its prospective nature is required for social interaction - shaping future behaviour based on prior experience...
The Memory-Centred Cognition perspective places an active association substrate at the heart of cognition, rather than as a passive adjunct. Consequently, it places prediction and priming on the basis of prior experience to be inherent and fundamental aspects of processing. Social interaction is taken here to minimally require contingent and co-adaptive behaviours from the interacting parties. In this contribution, I seek to show how the memory-centred cognition approach to cognitive architectures can provide an means of addressing these functions. A number of example implementations are briefly reviewed, particularly focusing on multi-modal alignment as a function of experience-based priming. While there is further refinement required to the theory, and implementations based thereon, this approach provides an interesting alternative perspective on the foundations of cognitive architectures to support robots engage in social interactions with humans.

\end{abstract}


% \fi
\IEEEpeerreviewmaketitle

%%%%%%%%%%%%%%%%%%%%%%%%%%%%%%%%%%%%%%%%%%%%%%%%%%%%%%%%%%%%%%%%%%%%%%%%%%%
%%%%%%%%%%%%%%%%%%%%%%%%%%%%%%%%%%%%%%%%%%%%%%%%%%%%%%%%%%%%%%%%%%%%%%%%%%%
%%%%%%%%%%%%%%%%%%%%%%%%%%%%%%%%%%%%%%%%%%%%%%%%%%%%%%%%%%%%%%%%%%%%%%%%%%%
\section{Introduction}

The representation and handling of memory is an important feature of cognitive architectures, with a variety of symbolic and sub-symbolic representation schemes used (generally as passive storage), typically based on assumptions of modularity \cite{Sun2004}. As such, memory is generally considered to be structurally separable from the cognitive processing mechanisms, and functions to provide these `cognitions' with the required data.

In the memory-centred cognition perspective, memory is instead considered to be a fundamentally active process that underlies cognitive processing itself rather than being a passive adjunct \cite{Baxter2011,Wood2012}. Based on evidence and models in neuropsychology, e.g. \cite{Fuster1997}, this approach necessitates a re-examination of the organisation and functions of cognitive architectures, as outlined below (section \ref{sec:arch}).

Previously, I put forward the case for the greater consideration of memory in HRI developments \cite{Baxter2014}. I argued that memory is pervasive: fundamentally involved in all aspects of social behaviour, beyond mere passive storage of information in data structures. In this brief (and relatively introspective) contribution, I expand on this point, exploring specifically the requirements of social interaction for robots, and consequently what cognitive architectures need to encompass.

%%%%%%%%%%%%%%%%%%%%%%%%%%%%%%%%%%%%%%%%%%%%%%%%%%%%%%%%%%%%%%%%%%%%%%%%%%%
%%%%%%%%%%%%%%%%%%%%%%%%%%%%%%%%%%%%%%%%%%%%%%%%%%%%%%%%%%%%%%%%%%%%%%%%%%%
%%%%%%%%%%%%%%%%%%%%%%%%%%%%%%%%%%%%%%%%%%%%%%%%%%%%%%%%%%%%%%%%%%%%%%%%%%%
\section{Facets of Social Interaction}\label{sec:social}

Social interaction is a complex phenomena that entails a range of abilities on the part of the interactants; indeed, there are facets of human-human social interaction that are as yet not fully understood, with the neural substrates supporting these in the individual yet to be characterised. One aspect that is commonly emphasised is the requirement for social signal processing for the individual, where behavioural cues (such as gaze, intonation, gesture, etc) should be interpreted to inform the behaviour of the observer.

One central idea emerging in the behavioural sciences is the notion of 'social contingency': the coupling and co-dependency of behaviours between interacting individuals \cite{DiPaolo2012}. This explicitly acknowledges the necessary role that the 'other' plays to set up the contingent behaviours, and moves away from the emphasis on social signal processing (though not discounting it). Minimal interaction paradigms provide intriguing illustrations of this: even given a low bandwidth interaction environment, there are non-trivial dynamics set up that cannot be explained by observations of an individual \cite{DiPaolo2008}. 

For social interaction generally, and in particular for this latter interacting systems perspective, there is an important role for prediction \cite{Brown2012}. When interacting, there is an expectation that the interaction partner is also a social agent, and thus predicable in that context. Infants, for example, can use the gaze behaviour of a robot to infer that the robot is a psychological agent with which they can interact \cite{Meltzoff2010}. A previous study has further lent support to the idea that the imposition of expectations of social behaviour (and therefore the arising of socially contingent behaviours, in this case turn-taking) will come about if the interactants view each other as (potentially) social agents \cite{Baxter2013a}. %In this case, it was the child viewing the robot as a potentially social agent, and thus engaging in turn-taking, even though the robot made no explicit allowance for the child behaviour.

If the interaction partner (whether it is human or robot) is attributed with social agency, initially as a result of anthropomorphism for example \cite{Duffy2003}, then one fundamental characteristic of social interaction between humans that will be seen is the `chameleon effect' \cite{Chartrand1999}, or imitation/alignment, e.g. \cite{Dautenhahn1999,Baxter2013,Vollmer2015}. The presence of this within an interaction, as a type of contingency between the interactants (see above), could be seen as an indicator of sociality.

These phenomena, from attribution of social agency to alignment, illustrate a necessity for social robots (to a certain extent at least) to conform to human cognitive and behavioural features, as well as to their constraints, to enable predictability, consistency and contingency of robot behaviour with respect to the human(s) in the interaction.

%%%%%%%%%%%%%%%%%%%%%%%%%%%%%%%%%%%%%%%%%%%%%%%%%%%%%%%%%%%%%%%%%%%%%%%%%%%
%%%%%%%%%%%%%%%%%%%%%%%%%%%%%%%%%%%%%%%%%%%%%%%%%%%%%%%%%%%%%%%%%%%%%%%%%%%
%%%%%%%%%%%%%%%%%%%%%%%%%%%%%%%%%%%%%%%%%%%%%%%%%%%%%%%%%%%%%%%%%%%%%%%%%%%
\section{Memory-Centred Cognitive Architecture}\label{sec:arch}

%Multi-modality processing, developmental structure and inherent prediction through experience, priming (as multi-modal prediction for example)... 

From neurospychology, the Network Memory framework \cite{Fuster1997} emphasises the central role that distributed associative cortical networks play in the organisation and implementation of cognitive processing in humans. The role of associative networks serves not only as a learning system (through Hebbian-like learning), but also as a substrate for activation dynamics. The reactivation and adaptation of existing networks combine to generate behaviour that is inherently based on prior experience.

The Memory-Centred Cognition perspective, as applied to the domain of cognitive robotics \cite{Baxter2011}, seeks to extend these principles of operation: associative networks supporting activation dynamics that bring prior experience to bear on the current situation. A developmental perspective is necessary in order to do so \cite{Baxter2010b}: the creation (and subsequent updating) of the associative networks must be done through the process of experience in order to form the appropriate associations between information in the present sensory and motor modalities of the robot (or system, in the case of a simulation).

Once an associative structure has been acquired, the principle mechanism at play is \textit{priming} \cite{Baxter2011}. Priming in a memory-centred system occurs when some sub-set of the system is stimulated (from incoming sensory information for example), which causes activation to flow around the network, in turn causing parts of the network with no external stimulation to become active. %For example, if there is a strong association between a specific person and a specific cup, then the presence of this person would activate the `representation' of the cup.
Priming in this way fulfils a number of important functions. Firstly, it sets up cross-modal expectations, or the prediction of currently absent stimuli. 
Secondly, the priming process facilitates an integration of information across different modalities in a way that is explicitly based on prior experience (biased by the weights of the associative network).

A computational implementation of this has been applied to an account of the developmental acquisition of concepts \cite{Baxter2012a}: not only was the system able to complete the task with a high success rate, but also the errors it made were consistent with those made by humans.
A similar computational implementation has also been used to demonstrate how word labels for real-world objects can facilitate further cognitive processing \cite{Morse2011}. These examples provide a glimpse of the range of cognitive processing (relevant to human cognitive processing) that can be accounted for using the memory-centred perspective.

Regarding social human-robot interaction, and in particular the notion that alignment is a fundamental feature of it (section \ref{sec:social}), the memory-centred perspective provides an intuitive, and indeed effective, account. Using exactly the same mechanism as for the concept learning study, the structure of an associative network was learned based on human behaviour (across a number of different modalities), which could then be directly used to determine the characteristics of the robot behaviour \cite{Baxter2013}. Alignment is achieved as a by-product of the way the memory-centred cognitive system operated: the associations were learned through experience, and behaviour was generated from priming (i.e. recall).

%%%%%%%%%%%%%%%%%%%%%%%%%%%%%%%%%%%%%%%%%%%%%%%%%%%%%%%%%%%%%%%%%%%%%%%%%%%
%%%%%%%%%%%%%%%%%%%%%%%%%%%%%%%%%%%%%%%%%%%%%%%%%%%%%%%%%%%%%%%%%%%%%%%%%%%
%%%%%%%%%%%%%%%%%%%%%%%%%%%%%%%%%%%%%%%%%%%%%%%%%%%%%%%%%%%%%%%%%%%%%%%%%%%
\section{Addressing Questions}

From the context outlined above, I now attempt to provide answers to a set of six questions relevant to the notion of social cognitive architectures.
%Partly given space constraints, but primarily due to the ongoing development of both my theoretical perspective and computational systems, these answers do not promote a particular set of algorithms, but rather a broader perspective. 
I particularly seek to emphasise a principled-basis (as opposed to computational mechanism-basis) for cognitive architectures and for the application to social interaction. 
%If it is a successfully implemented and validated social robot you want, then you will not find it detailed below; rather, I hope you will find my approach outlined to eventually attaining this goal. 

\subsection{Why should you use cognitive architectures - how would they benefit your research as a theoretical framework, a tool and/or a methodology?}\label{sec:1}

The benefit would be in considering cognitive architectures as a set of principles (a theoretical framework), a methodology for assessing these principles, and as a tool for providing robots with autonomous intelligent behaviour.

There are in my view three specific contributions related to scientific development (as opposed to technical implementation) that cognitive architectures can make to HRI research and development, which are centred around the idea of a cognitive architecture being made up of a set of formalised hypotheses. 

Firstly, in a principled manner, they allow data and theory from empirical human studies to be integrated into artificial systems. For example, if data from a psychology experiment is to be integrated, a framework for doing so is required (i.e. the architecture enables an interpretation of the data). This first point promotes the idea of a directly human-inspired/constrained architecture.
Secondly, treating cognitive architectures as a set of formalised (through implementation) principles, they facilitate a comparison of different architectures at a level abstracted away from the computational systems/algorithms used, enabling a focus on the assumptions. In the presently considered case of social interaction, this is a useful facet given the as yet uncertain nature of what exactly constitutes social interaction (section \ref{sec:social}).
Thirdly, the application of cognitive architectures (in robotic systems for instance) provides a means of evaluating its constituent assumptions and principles.
This is related to the first point, but is focused more on the integration of empirical evidence obtained from application/experimentation with the architecture itself.

\subsection{Should cognitive architectures for social interaction be inspired and/or limited by models of human cognition?}\label{sec:2}

Following from the principles of social interaction outlined above, essentially, yes.

Taking the view that social interaction between humans is founded on the intrinsic tendency of humans to expect certain types of behaviour from their interaction partners (see section \ref{sec:social}), it becomes important to ensure that the robot will not violate expectations. In order not to violate expectation, there must necessarily be some understanding (either on the part of the system designer or learned by the system itself) of what expected human behaviour would be. 

In the memory-centred cognition perspective, prior interaction history of the robot with humans would constrain its future behaviour by this experienced behaviour.

\subsection{What are the functional requirements for a cognitive architecture to support social interaction?}\label{sec:3}

%Need to provide a (working) definition of social interaction. Anticipation, contingent reactions, user modelling (superficial/deep?), consideration of all aspects of the agent, not just the cognitive architecture (example of keepon \cite{Kozima2006,Peca2015} as being very effective)...

The discussion of social interaction (section \ref{sec:social}) emphasised the importance of contingent behaviour, anticipation/prediction to support this, and adaptation/personalisation. In addition, it is necessary to specify appropriate timing, and embodiment-appropriate responses.

If socially-appropriate behaviour is in the eye of the (human) beholder, then the Keepon robot for example demonstrates the importance of coherence of behaviour and timing \cite{Kozima2006}. The minimally complex embodiment is convincingly responsive in a social manner, to the extent that it is seen as a communicative partner \cite{Peca2015}. Even though it doesn't use language, only uses few degrees of freedom (in contrast to many other robots used in HRI), and is only minimally humanoid in appearance, the effect of apparent sociality is strong.

Integration of sensory and motor modalities in a temporally consistent and responsive manner (i.e. contingency), based on principles of prediction from prior experience (i.e. memory), and coherency with the robot embodiment used (c.f. Keepon example) are therefore fundamental functional requirements for a social cognitive architecture.

\subsection{How would the requirements for social interaction inform your choice of the fundamental computational structures of the architecture (e.g. symbolic, sub-symbolic, hybrid, ...)?}\label{sec:4}

%The inherent advantages that sub-symbolic representation schemes afford - however, the drawback that at the moment, there is no complete account of how this may happen in only sub-symbolic architectures - the symbolic element is as yet still required for the requisite functionality. My aim however is to keep pushing 'up' the boundary between symbolic and sub-symbolic...

Given the commitment to the memory-centred cognition perspective in this work, there is a natural fit with sub-symbolic computational structures. This provides a number of inherent advantages (section \ref{sec:arch}), such as the integration of predictive behaviour from prior experience, and priming effects (within and between modalities).

However, the nature of applications in human-robot interaction (relying on language for example) means that it is not yet possible to dispense with symbol-processing systems. Nevertheless, there is in principle an effort to push the limits of sub-symbolic processing mechanisms up the processing and representation hierarchy, as revisited below (section \ref{sec:discussion}).

\subsection{What is the primary outstanding challenge in developing and/or applying cognitive architectures to social HRI systems?}\label{sec:5}

%Limits in understanding of what is actually involved for humans, their expectations etc. Also, the technical challenges of rich sensory data acquisition and interpretation.

One of the primary challenges in the application of cognitive architectures to social interaction lies in the general lack of understanding of what is precisely involved in human-human social interaction. To a certain extent it is an attempt to find a solution to a problem that is as yet not fully characterised. This reflects on the requirements for the cognitive architectures that should engage in social interaction: if a commitment to human-like cognition/behaviour is made (see section \ref{sec:2}), then what precisely are the constraints that need to be incorporated?

A more practical concern that requires further development is the provision of sensory systems for robots that can provide sufficiently complex characterisations of the (social) environment for effective decision making. There is however, in my opinion, no clear distinction between sensory systems and cognitive processing, given the necessity for interpretation of raw sensory signals (e.g. camera images) at various levels of abstraction.

\subsection{Can you devise a social interaction scenario that current cognitive architectures would likely fail, and why?}\label{sec:6}

%Maybe... Long-term interaction?

%Given a single benchmark interaction scenario, it seems likely that there would be a solution found using current architectures and technologies. 
The question is whether the application to a single domain can be generalised to other domains, which is where the benefits of cognitive architectures should come (section \ref{sec:1}).
As such, rather than a specific interaction scenario, I would suggest instead that autonomous sociality over variable time-scales poses challenges to current approaches and implementations.

In the short term, the challenge for social robots is to produce behaviour appropriate to the interaction context, informed by prior interaction experience, in a manner consistent with the expectations of the interacting humans. Furthermore, this socially interactive behaviour should adapt to the interaction partner over time, in terms of verbal and non-verbal behaviours for example. The technical challenges to support this in terms of sensory processing are outstanding, but there are also clear challenges in terms of the mechanisms of adaptation required (i.e. the `cognitive' aspect). The memory-centred approach has ventured an implementation towards this problem, although the account is as yet incomplete.

Over extended periods of time, the challenges are compounded by requirements for stability. This is not just stability in terms of ensuring the system doesn't fail, but also in resolving the apparent trade-off between adaptability to new situations and robustness of the cognitive system. From the perspective of the memory-centred cognition account, the resolution to this question lies in how the formation, maintenance and manipulation of memory is handled in the system in terms of parameters and structures.

%%%%%%%%%%%%%%%%%%%%%%%%%%%%%%%%%%%%%%%%%%%%%%%%%%%%%%%%%%%%%%%%%%%%%%%%%%%
%%%%%%%%%%%%%%%%%%%%%%%%%%%%%%%%%%%%%%%%%%%%%%%%%%%%%%%%%%%%%%%%%%%%%%%%%%%
%%%%%%%%%%%%%%%%%%%%%%%%%%%%%%%%%%%%%%%%%%%%%%%%%%%%%%%%%%%%%%%%%%%%%%%%%%%
\section{Outlook}\label{sec:discussion}

%In summary...

The nature of the discussion above is primarily principled and theoretical rather than focused on specific computational mechanisms. Naturally I believe memory-centred cognition perspective to have a consistency and coherence that merits consideration and further development. However, it is not in its current state able to practically support all aspects of real social interactions with real people.

This is a limitation shared with many `emergent' cognitive architecture approaches \cite{Vernon2007}: theoretically interesting and coherent perhaps, but practically limited in terms of what can be done on real systems (use of language and dialogue being good examples of this). This is partly due to an implication of the theoretical perspective: by committing to a holistic approach that emphasises the integration and interplay of many different factors (including, for example, cognition, embodiment, culture, etc), the problem is made more difficult before a computational implementation is even begun. On a practical level, the types of dynamical system (be they neural network-based or other) used are typically not fully understood, or are at least highly complex \cite{Beer1995}, e.g. in terms of conditions for stability (particularly when adaptation/learning is incorporated), which does not bode well for social robots that have to be reliable in real interactions with real people.

For these reasons, I do not believe that symbol-based approaches should (or can) be discarded, at least not for the foreseeable future. They provide the means of getting closer to actually achieving the desired behaviours in reality. Having said this, and as noted above (sec. \ref{sec:4}), I remain intent on pushing the boundary between symbolic and sub-symbolic implementations `up' the abstraction hierarchy, in a manner common with a range of other developmentally-oriented researchers \cite{Smith2005,Cangelosi2010}.%

So, what does a memory-centred cognitive architecture look like if it is to be effectively applied to social interaction? And what does the memory-centred cognitive architecture enable in terms of social robots that would be difficult to achieve with an alternative approach? The functionality of developmental learning of cross-modal associations for prediction and action generation outlined above (section \ref{sec:arch}) provides a technically difficult but in principle effective solution to the issue of learning from a vast array of potential multi-modal information in a way that is useful for action generation.
This is not to say that this is the only approach (theoretical or computational) that would be capable of a similar functionality.
However, this is where the second aspect, the requirement to fulfill social interaction with humans through conformity with human cognition (section \ref{sec:social}), becomes a distinguishing characteristic of the memory-centred approach.

In developing the theory, I have applied it to a range of practical systems and applications, as reviewed above (section \ref{sec:arch}). For example using the same mechanism, accounts have been made of concept acquisition \cite{Baxter2012a} and multi-modal robot behaviour alignment to an interaction partner \cite{Baxter2013}. Other systems using the same principles have been used to demonstrate the development of low-level sensory-motor coordination through experience \cite{Baxter2010b}, and the role of words in supporting new cognitive capabilities \cite{Morse2011}.

Whereas my commitment to the memory-centred cognition perspective for robotics is strong, my commitment to the specific mechanisms used is weak. I must acknowledge that there are a number of weaknesses with the various systems used, notably related to hierarchical structure/representation, and an incomplete account of temporal processing.
However, in my view, this does not invalidate the theoretical approach, and merely serves to provide motivation to either find or develop a more appropriate computational implementation that fulfils all of the principles and constraints of the memory-centred cognition perspective.

%%%%%%%%%%%%%%%%%%%%%%%%%%%%%%%%%%%%%%%%%%%%%%%%%%%%%%%%%%%%%%%%%%%%%%%%%%%
%%%%%%%%%%%%%%%%%%%%%%%%%%%%%%%%%%%%%%%%%%%%%%%%%%%%%%%%%%%%%%%%%%%%%%%%%%%
%%%%%%%%%%%%%%%%%%%%%%%%%%%%%%%%%%%%%%%%%%%%%%%%%%%%%%%%%%%%%%%%%%%%%%%%%%%
\section*{Acknowledgement}

This work was supported by the EU FP7 project DREAM (grant number 611391, http://dream2020.eu/).

%%%%%%%%%%%%%%%%%%%%%%%%%%%%%%%%%%%%%%%%%%%%%%%%%%%%%%%%%%%%%%%%%%%%%%%%%%%
%%%%%%%%%%%%%%%%%%%%%%%%%%%%%%%%%%%%%%%%%%%%%%%%%%%%%%%%%%%%%%%%%%%%%%%%%%%
%%%%%%%%%%%%%%%%%%%%%%%%%%%%%%%%%%%%%%%%%%%%%%%%%%%%%%%%%%%%%%%%%%%%%%%%%%%
%\begin{thebibliography}{1}

%\end{thebibliography}

\bibliographystyle{IEEEtran}
\bibliography{Bib}

\end{document}